\documentclass{article}

\title{Using Answer Set Programming for pattern mining\blfootnote{This work first appeared in the proceedings of the IAF conference (French Fundamental Artificial Intelligence Conference) in june 2014.}}

\author{Thomas Guyet$^1$, Yves Moinard$^2$ and Ren\'e Quiniou$^2$ \vspace{.5cm} \\
\small $^1$ AGROCAMPUS-OUEST/IRISA UMR6074, 35042 Rennes, France\\
\small $^2$ \textsc{Inria}, Campus de Beaulieu, 35042 Rennes, France
}

\date{}

\newcommand\blfootnote[1]{%
  \begingroup
  \renewcommand\thefootnote{}\footnote{#1}%
  \addtocounter{footnote}{-1}%
  \endgroup
}

\usepackage{graphicx}
\usepackage{amsmath, amssymb}
\usepackage{cancel}
\usepackage[ruled, vlined]{algorithm2e}
\usepackage{hyperref}

\newcommand{\ie}{\emph{i.e.~}}
\newcommand{\cf}{\emph{cf.~}}
\newcommand{\eg}{\emph{e.g.~}}
\newcommand{\al}{\emph{al.~}}
\newcommand{\espace}{\vspace{10pt}}

\usepackage{amsthm}

\newtheorem{example}{Example}
\newtheorem{definition}{Definition}
\newtheorem{property}{Property}

\usepackage{listings}
\begin{document}

\maketitle


\subsubsection*{Abstract}
Serial pattern mining consists in extracting the frequent sequential patterns from a unique sequence of itemsets. This paper explores the ability of a declarative language, such as Answer Set Programming (ASP), to solve this issue efficiently.
We propose several ASP implementations of the frequent sequential pattern mining task: a non-incremental and an incremental resolution.
The results show that the incremental resolution is more efficient than the non-incremental one, but both ASP programs are less efficient than dedicated algorithms. Nonetheless, this approach can be seen as a first step toward a generic framework for sequential pattern mining with constraints.
%
\section{Introduction}

Sequential pattern mining aims at analysing ordered or timed data to extract interesting patterns the elements. 
Broadly, a pattern is considered interesting if it occurs frequently in the data, \ie the number of its occurrences is greater than a fixed given threshold.

As non informed mining methods tend to generate massive results, there is more and more interest in pattern mining algorithms able to mine data considering some expert knowledge.
Though a generic pattern mining tool that could be tailored to the specific task of a data-scientist is still a holy grail for pattern mining software designers, some recent attempts have proposed generic pattern mining tools \cite{guns2013miningzinc,negrevergne2013dominance} for itemset mining tasks.

One way to introduce expert knowledge in pattern mining algorithms is to define constraints on the expected patterns. Motivated by the use of constraints in sequential pattern mining. Pei et \al \cite{pei2002mining,pei2007seqmining} defined seven types of generic constraints that a data scientist would like to express.
However, beyond generic constraints, a data scientist would like to express a lot of domain-specific constraints reflecting his own knowledge and preferences.


ASP (Answer Set Programming) is a declarative language that enables an easy expression of constraints as well as domain knowledge and associated formal reasoning tools. To engage in the generic way sketched before, this paper proposes to implement a classical sequential pattern mining algorithm in ASP with the future objective of letting the data scientist define constraints on its patterns and associated domain knowledge. 

We show how to encode sequential pattern mining tasks in ASP. The first intuitive and naive encoding is next improved with the incremental resolution facilities of the \texttt{clingo} ASP solver \cite{gekakaosscsc11a}. The performance of the method is evaluated on simulated data and compared with the dedicated algorithm MinEpi \cite{Mannila1997}.

\section{State of the art}

Designing pattern mining languages based on pattern constraints has developed recent interest in the literature \cite{guns2013miningzinc,negrevergne2013dominance,Bonchi:2007:ESC:1224242.1224353,Boulicault2005}. The aim of these proposals is to obtain a \textit{declarative} constraint-based language even at the cost of degraded runtime performance compared to a specialized algorithm.


All these approaches have been conducted on itemset mining in transaction databases, which is much simpler than sequential pattern mining in a sequence database, which in turn is simpler than sequential pattern mining in a unique long sequence.

Sequential pattern mining in a sequence database have been addressed by numerous algorithms inspired by algorithms for mining frequent itemsets. The most known algorithms are GSP \cite{Srikant96}, SPIRIT \cite{SPIRIT1999}, SPADE \cite{Zaki01}, PrefixSpan \cite{Pei2004}, and CloSpan \cite{CloSpan03} or BIDE \cite{BIDE2004} for closed sequential patterns.
To determine the support of a pattern, these algorithms count the number of sequences in the database that contain the pattern without taking into account repetitions of the pattern in the sequences. Deciding if a sequence contains a pattern is generally simple.
It is worth-noting that all these algorithms satisfy the anti-monotonicity property which is essential to obtain good mining performances. The anti-monotonicity property says that if some pattern is frequent then all its sub-patterns are also frequent. And reciprocally, if some pattern is not-frequent then all its super-patterns are non-frequent. This property enables the algorithm to prune efficiently the search space and thus reduces its exploration.

Counting the number of occurrences of a pattern, or of patterns, in a long sequence introduces some complexity compared to previous approaches as two occurrences of a pattern can overlap. Defining how to enumerate the occurrences of a pattern is important to ensure the anti-monotonicity property.
The MinEpi and WinEpi algorithms \cite{Mannila1997} extract frequent \textit{episodes}. An episode is a directed acyclic graph where edges express the temporal relations between events. A \textit{proper edge} from node $v$ to node $w$ in the pattern graph implies that the events of $v$ must occur before the events of $w$, while a \textit{weak edge} from $v$ to $w$ implies that the events of $w$ may occur either at the same time as those of $v$ or later. Testing whether an episode occurs in the sequence is an NP-complete problem \cite{Tatti2011a}. 

The frequency measure used by MinEpi relies on minimal occurrences of patterns and is anti-monotonic.
Other counting methods have been proposed to solve similar problems while preserving the anti-monotonicity property required for the effectiveness of pattern occurrences search (see \cite{Achar2010} for a unified formulation of sequential pattern counting methods).

\espace

Integrating constraints in sequential pattern mining is often limited to the use of simple anti-monotonic temporal constraints such as \texttt{mingap} constraints. In addition to the classical support constraint, Pei et \al \cite{pei2002mining} defined seven types of constraints on patterns. SPIRIT \cite{SPIRIT1999} is one of the rare algorithm that considers complex constraints on patterns such as regular expressions.

In \cite{guns2011CPmining}, Guns et \al proposed a new perspective on data mining based on constraint programming. Their aim is to separate modelling and solving in data mining in order to gain more flexibility for elaborating new mining models. They chose to illustrate the approach on itemset mining. Only few recent works have proposed to explore constraint programming for sequential pattern mining. M\'etivier, Loudni et \al \cite{Metivier2013} have developed a constraint programming method for mining sequential patterns with constraints in a sequence database. The constraints are based on \textit{amongst} and \textit{regular expression} constraints and expressed by automata.
Coquery et \al \cite{coquery2012sat} have proposed a SAT based approach for sequential pattern mining. The patterns are of the form $ab?c$ and an occurrence corresponds to an exact substring (without gap) with joker (the character $?$ replaces exactly one item different from $b$ and $c$).

\espace

To the best of our knowledge, J\"arvisalo's work \cite{jarvisalo2011itemset} is the unique application of ASP to pattern mining. Following Guns et \al's proposal, J\"arvisalo designed an ASP program to extract frequent itemsets in a transaction database. A main feature of J\"arvisalo is that each answer set (AS) contains only one frequent itemset associated with the identifiers of the transactions where it occurs.
J\"arvisalo addressed this problem as a new challenge for the ASP solver, but did not highlight the potential benefit of this approach to improve the expressiveness of pattern mining tools.

\espace

In this paper, we explore the use of ASP to extract frequent string patterns (\ie patterns without parallel events) in a unique long sequence of itemsets where the occurrences of a string pattern are the minimal occurrences. In addition, we implemented the following simple anti-monotonic constraints to improve the algorithm efficiency:
\begin{enumerate}
\item{an item constraint} states which particular items should not be present in the patterns.
\item{a length constraint} fixes the maximal length of patterns (number of items).
\item{a duration constraint} sets the maximal duration of patterns (time elapsed between the occurrence of the first itemset and the occurrence of the last one - difference between timestamps).
\item{a gap constraint} sets the maximal time delay between the respective timestamps of two successive itemsets.
\end{enumerate}

\section{Sequential pattern mining} \label{sec:Notations}

\subsection{Items, itemsets and sequences}
From now on, $[n]$ denotes the set of the $n$ first integers, \ie $ [n]= \{1, \dots, n \} $.

Let $\mathcal{E}$ be the set of items and $\eqslantless$ a total (reflexive) order on this set (\eg lexicographical). An \emph{itemset} $A=(a^1, a^2, \ldots, a^n),\; a^i\in\mathcal{E}$ is an ordered set of distinct items, \ie $\forall i \in [n-1],\; a^i \eqslantless a^{i+1}$ and $i \neq j \Rightarrow a^i\neq a^j$.
The size of an itemset $\alpha$, denoted $|\alpha|$ is the number of items it contains.
An itemset $\beta=(b^1, \ldots, b^m)$ is a sub-itemset of $\alpha=(a^1, \ldots, a^n)$, denoted $\beta \sqsubseteq \alpha$, iff $\beta$ is a subset of $\alpha$.

A \emph{sequence} $S = \langle s_1, s_2, \ldots, s_n \rangle$ is an ordered series of itemsets $s_i$. 
The length of a sequence $S$, denoted $|S|$, is the number of itemsets that make up the sequence. The size of a sequence $S$, denoted $\|S\|$, is the total number of items it contains:~ $\|S\| = \sum_{i=1}^{|S|}|s_i|$.

$T=\langle t_1, t_2, \ldots, t_m\rangle$ is a \emph{sub-sequence} of $S=\langle s_1,$ $s_2,$ $\ldots, s_n\rangle$, denoted $T \preceq S$, iff there exists a sequence of integers $1 \leq i_1 < i_2 < \ldots  < i_m \leq n$ such that $\forall k \in [m], t_k \sqsubseteq s_{i_k}$.
$T=\langle t_1, t_2, \ldots, t_m\rangle$ is a \emph{prefix} of $S=\langle s_1, s_2, \ldots, s_n\rangle$, denoted $T \preceq_b S$\footnote{The $b$ in $\preceq_b$ stands for \emph{b}ackward subsequence.}, iff $\forall k \in [m-1], t_k = s_{k}$ and $t_m \sqsubseteq s_{m}$, $t_m \neq s_m$.

\begin{example}
Let $\mathcal{E}=\{a,b,c\}$ with the lexicographical order ($a\eqslantless b$, $b\eqslantless c$) and the sequence $S=\langle a(bc)c(abc)cb\rangle$. To simplify the notation, we omit the parentheses around itemsets containing a single item, and comas inside itemsets. 
The size of $S$ is 9 and its length is 6. For instance, sequence $\langle (bc)(ac)\rangle$ is a sub-sequence of $S$ and $\langle a(bc)c(ac)\rangle$ is a prefix of $S$.
\end{example}

The relations $\preceq$ and $\preceq_b$ define partial orders on sets of sequences.

\subsection{Mining string patterns in a long sequence}

A \emph{long sequence} $F=\{f_i\}_{i\in \mathbb{N}}$ is a single sequence of itemsets. A \textbf{string pattern}  $P=\langle p_1, \ldots, p_n \rangle$ is a sequence of items $(p_k)$ (all itemsets are singleton). In the sequel, we use the term \textbf{pattern} to refer to our notion of string pattern. Our notion of pattern is more specific than the one defined by Mannila et \al \cite{Mannila1997}.

\begin{definition}[Occurrences of a pattern, minimal occurrences]
\label{def:Instances}
An occurrence of the pattern $P=\langle p_1, \ldots, p_n \rangle$ of length $n$ in a long sequence $S=(s_1, \ldots, s_m)$ is noted by the $n$-tuple $T=(t_1,  \ldots, t_n)$, where $t_k, k \in [n]$ is the position of the occurrence of element $p_k \in s_{t_k}$ in $S$. $P$ is said to occur at interval $[t_1, t_n]$ in $S$.

Let $T=(t_1,  \ldots, t_n)$ and $T'=(t'_1,  \ldots, t'_n)$, 
$$T \lhd T' \Leftrightarrow \left\{\begin{array}{l}
[t_1,t_n] \subset [t'_1, t'_n], \text{or }\\
 n>1 \wedge  t_1=t'_1 \wedge t_n=t'_n \wedge \\ \:\:\:\: (t_1,  \ldots, t_{n-1})\lhd (t'_1,  \ldots, t'_{n-1})
\end{array}\right.$$

An occurrence $T$ of a pattern $P$ is minimal if it is minimal for $\lhd$.
$\mathcal{I}_{S}(P)$ denotes the set of all minimal occurrences of $P$ in $S$.
\end{definition}

\begin{example} \label{ex:minimaloccurrences}
Let $S = \langle a (bc) (ac)c d \rangle $ be a long sequence and $P=\langle abc\rangle$ be a pattern. Finding occurrences of pattern $P$ in $S$ means to locate items of $P$ in $S$: $\langle a\rangle$ appears only at positions 1 and 3, $\langle b \rangle $ appears at position 2, $\langle c\rangle$ appears at positions 2, 3 and 4. Pattern $P$ could has two occurrences: $(1,2,3)$ and $(1, 2, 4)$. The minimal occurrence condition eliminates occurrence $(1,2,4)$ because $[1,2]$ is (strictly) included in $[1,4]$. Thus $\mathcal{I}_{S}(\langle abc\rangle)=\{(1,2,3)\}$.

We now consider the pattern $P=\langle acd\rangle$ in order to illustrate the second (recursive) minimal occurrence condition. The occurrences of the pattern would be $\{(1,2,5), (1,3,5)\}$. The occurrence $(1,3,5)$ is not minimal because the occurrences bounds are equivalents but $[1,2]$ is (strictly) included in $[1,3]$, thus $(1,2,5)\lhd (1,3,5)$.

Let $S=\langle aaaa\rangle$ be the long sequence. $\mathcal{I}_{S}(\langle aa\rangle)=\{(1,2), (2,3), (3,4)\}$ and $\mathcal{I}_{S}(\langle aaa\rangle)=\{(1,3), (2,4)\}$.
\end{example}

Let $S$ be a long sequence and $P$ a pattern. The \textbf{support} of pattern $P$, denoted by $supp(P)$, is the cardinality of $\mathcal{I}_{S}(P)$, \ie $supp(P) = card\left(\mathcal{I}_{S}\left(P\right)\right)$.

Note that the support function $supp(\cdotp)$ is not anti-monotonic on the set of patterns with associated partial order $\preceq$ (see \cite{Tatti2012}). Considering order $\preceq_b$, $supp(\cdotp)$ is anti-monotonic.

\begin{definition}[Mining a long sequence]
Given a threshold $\sigma$, a pattern $P$ is \emph{frequent} in a long sequence $S$  iff $supp(P) \geq \sigma$. Mining a long sequence consists in extracting all frequent patterns.
\end{definition}

\begin{example}
Let $\sigma=2$. The set of frequent patterns in 
$S = \langle a (bc) (abc) c (bc) \rangle $ is \{
$\langle a\rangle$, 
$\langle b\rangle$, 
$\langle c\rangle$, 
$\langle ab\rangle$, 
$\langle ac\rangle$, 
$\langle bb\rangle$, 
$\langle bc\rangle$,
$\langle cb\rangle$, 
$\langle cc\rangle$, 
$\langle acb\rangle$, 
$\langle acc\rangle$
$\langle bcc\rangle$
$\langle ccc\rangle$
\}.
\end{example}

\section{Mining serial patterns with ASP} \label{sec:miningasp1}

This section shows a first way of specifying serial pattern mining in classical ASP programming style. After presenting how to model the data in ASP, we give an intuitive algorithm, then we show how to improve its resolution using \texttt{clingo} control abilities.

Our proposal is borrowed from J\"arvisalo's \cite{jarvisalo2011itemset}: a solution of the ASP program is an answer set (AS) that contains a single frequent pattern as well as its occurrences. The resolution relies on the ``generate and test principle'': generate combinatorially all the possible patterns, one at a time, associated with their minimal occurrences and test whether they satisfy the specified constraints.

\subsection{Modelling the long sequence, patterns and occurrences}

A long sequence is modelled by the predicate \lstinline!seq/2!. For example, the atom \lstinline!seq(3,5)! declares that an item $5$ occurs at timestamp $3$. Similarly, the current pattern is modelled by the predicate \lstinline!pattern/2!.

\begin{example}
Let $S = \langle a c a (ac) bc \rangle $ be a long sequence and $P=\langle acc\rangle$ be a pattern. 

\begin{lstlisting}[firstnumber=1]
%sequence description : a=1, b=2, c=3
seq(1,1). seq(2,3). seq(3,1). seq(4,1). seq(4,3). seq(5,2). seq(6,3).
%pattern description
pattern(1,1). pattern(2,3). pattern(3,3).
\end{lstlisting}
\end{example}

An occurrence of some pattern $P$ is described by a set of atoms \lstinline!occ(I,P,S)! where \lstinline!I! is the identifier of the occurrence, (an AS describes all the occurrences of its pattern),
 \lstinline!P! is the item position in the pattern and \lstinline!S! is the timestamp of a matching itemset in the sequence.

\begin{example}
Continuing example above, the occurrences of pattern $P$ are $\mathcal{I}_{S}(P)=\{(1,2,4), (3,4,6)\}$.
The AS for the program above should contain:
\begin{lstlisting}[numbers=none]
%occurrences
occ(1,1,1). occ(1,2,2). occ(1,3,4).
occ(2,1,3). occ(2,2,4). occ(2,3,6).
\end{lstlisting}
\end{example}

\subsection{An ASP program for extracting frequent patterns of length $n$}

In this section, we detail an ASP program that generates all the frequent patterns of length $n$. According to the ASP programming principle, the program contains a generation part and a constraint part. The generation part specifies the pattern structure as a set of \lstinline!pattern! atoms and their related occurrences as a set of \lstinline!occ! atoms. The constraints eliminate pattern candidates that are not frequent.

First, we introduce some additional predicates and constants:
\begin{itemize}
\item constant \lstinline!th! represents the minimum frequency threshold,
\item predicate \lstinline!patlen/1! sets the length of patterns,
\item predicate \lstinline!symb/1! specifies the items that can appear in a pattern.
\end{itemize}

\subsubsection{Generating patterns and occurrences}

The program below 
gives the rules for generating occurrences of patterns and their occurrences.
The rule in line 6 generates patterns containing exactly $L$ \lstinline!pattern! atoms, expressed combinatorially on the vocabulary specified by \lstinline!symb/1! atoms. 
As only minimal models are solutions of ASP programs, every AS will contain \lstinline!L! \lstinline!pattern! atoms defining only one pattern.

Next, at line 9, predicate \lstinline!instid/1! is used to enumerate exactly \lstinline!th! occurrences of the specified pattern. All the occurrences of a pattern could be generated, but only \lstinline!th! are kept in the solution, \ie the minimal number of occurrences for a pattern to be frequent. 

Finally, at line 12, for each of the \lstinline!th! pattern occurrences, a set of exactly \lstinline!L! \lstinline!occ! atoms is generated. In this combinatorial step, every possible association between a pattern item and a sequence timestamp \lstinline!P! is blindly considered.

\begin{lstlisting}[label=asp1:generation,firstnumber=5]
%generating all the patterns
L { pattern(1..L, P) : symb(P) } L :- patlen(L).

%list of occurrences
instid(1..th).

% I: occurrence id, L: pattern length,P: sequence position
L{ occ(I, 1..L, P): seq(P, S) }L :- instid(I), patlen(L).
\end{lstlisting}

\subsubsection{Minimal occurrences constraints}

The loose specification above generates many solutions, some being correct but redundant, \eg with different occurrence identifiers, some being incorrect, \eg with non minimal occurrences. Now, we add some constraints to keep only the required models.

The first constraint expresses that the item at position \lstinline!N! in the pattern must correspond to one of the items at timestamp \lstinline!P! in the given sequence. Constraints are expressed by negations. The following constraint requires that for any occurrence atom \lstinline!occ(I,N,P)! of a pattern having item \lstinline!S! at position \lstinline!N!, it is impossible to not have the same item \lstinline!S! at  timestamp \lstinline!P! in the sequence.

\begin{lstlisting}[firstnumber=13]
:- occ(I, N, P), pattern(N, S), not seq(P, S).
\end{lstlisting}

The next constraint expresses that the occurrences must respect the ordering of pattern items: it is not possible to have 2 items ordered in some way in the pattern and the 2 items they are mapped with ordered in the other way in the sequence.

\begin{lstlisting}[firstnumber=14]
:- occ(I, N, P), occ(I, M, Q), N<M, P>=Q.
\end{lstlisting}

The next constraint imposes that the $N$-th item of the pattern maps with a single timestamp in the sequence.
\begin{lstlisting}[firstnumber=15]
:- occ(I, N, P), occ(I, N, Q), P<Q.
\end{lstlisting}

Notice that here \lstinline!P < Q! is equivalent to \lstinline!P!!\lstinline!=Q!
 (meaning $P \neq Q$ in ASP) since \lstinline!P! and \lstinline!Q! play symmetric roles in the two atoms in \lstinline!occ!. This ``trick'' is often used in ASP (\cf also lines 16 and 17 in next listing), since it has the advantage of reducing the size of the grounding of the program.

The final constraints are related to minimal occurrences. The first constraint forbids solutions with two occurrences starting at the same timestamp. The second constraint forbids solutions with two occurrences ending at the same timestamp. The third constraint forbids solutions with two occurrences such that one ``contains'' the other.
\begin{lstlisting}[firstnumber=16]
:- occ(I, 1, P), occ(J, 1, Q), P=Q, I<J.
:- occ(I, L, P), occ(J, L, Q), patlen(L), P=Q, I<J.
:- occ(I, 1, SI), occ(J, 1, SJ), occ(I, L, EI), occ(J, L, EJ), patlen(L), SI<SJ, EJ<EI, I!=J.
\end{lstlisting}

Constraints 16 to 18 could be summed up by the single constraint below. However, the single constraint is significantly less computationally efficient than the three elementary constraints.
\begin{lstlisting}[numbers=none]
:- occ(I, 1, SI), occ(J, 1, SJ), occ(I, L, EI), occ(J, L, EJ), patlen(L), SI<=SJ, EJ<=EI, I!=J.
\end{lstlisting}

\subsubsection{Reducing the combinatorics of the solution space}

At this step, all the ASs are correct answers to the sequence mining problem, but many of them represents the same solution due to the combinatorial nature of the generation step. For instance, extracting patterns of length 4 in a sequence of 20 items yields 2448 answer sets, but there are only 24 different answer sets\footnote{This can be easily tested using the \texttt{project} option of \texttt{clingo}.}.

To improve the program efficiency, we add the following constraint on occurrence identifiers imposing that occurrences are ordered by their beginning timestamps. This constraint reduces the number of answer sets to 408. It is better but not perfect.

\begin{lstlisting}[firstnumber=19]
:- occ(I, 1, P), occ(J, 1, Q), I<J, P>Q.
\end{lstlisting}

\subsection{Illustration of constraint effectiveness}

Let $W=\langle abab\rangle$ be a sequence and $th=2$ the support threshold. 
We consider only the models generated with \lstinline!pattern(1,1)! and \lstinline!pattern(2,2)!, \ie that evaluates the occurrences of $\langle ab\rangle$. The models that contain at least two occurrences of $\langle ab\rangle$ contain atoms of the form $occ(1,w,\alpha)$, $occ(1,x,\beta)$ for occurrence $1$ and $occ(2,y,\gamma)$, $occ(2,z,\delta)$ for occurrence $2$, with $\alpha, \beta, \gamma, \delta \in [4]$ and $w, x, y, z \in [2]$.

Taking into account the unicity constraint, there are 256 models of the form $occ(1,1,\alpha)$, $occ(1,2,\beta)$, $occ(2,1,\gamma)$,  $occ(2,2,\delta)$, with $\alpha, \beta, \gamma, \delta \in [4]$.
Adding the mapping constraint, the valid models are the following (the 4-tuples below give the related values for $\alpha, \beta, \gamma, \delta$):

\begin{center}
\begin{tabular}{cccc}
$(1,2,1,2)$ & $(1,4,1,2)$ & $(1,2,1,4)$ & $(1,4,1,4)$ \\
$(3,2,1,2)$ & $(3,4,1,2)$ & $(3,2,1,4)$ & $(3,4,1,4)$ \\
$(1,2,3,2)$ & $(1,4,3,2)$ & $(1,2,3,4)$ & $(1,4,3,4)$ \\
$(3,2,3,2)$ & $(3,4,3,2)$ & $(3,2,3,4)$ & $(3,4,3,4)$ \\
\end{tabular}
\end{center}

The constraint stating that two occurrences cannot start on the same timestamp, eliminates 8 models:

\begin{center}
\begin{tabular}{cccc}
$\cancel{(1,2,1,2)}$ & $\cancel{(1,4,1,2)}$ & $\cancel{(1,2,1,4)}$ & $\cancel{(1,4,1,4)}$ \\
$(3,2,1,2)$ & $(3,4,1,2)$ & $(3,2,1,4)$ & $(3,4,1,4)$ \\
$(1,2,3,2)$ & $(1,4,3,2)$ & $(1,2,3,4)$ & $(1,4,3,4)$ \\
$\cancel{(3,2,3,2)}$ & $\cancel{(3,4,3,2)}$ & $\cancel{(3,2,3,4)}$ & $\cancel{(3,4,3,4)}$ \\
\end{tabular}
\end{center}

The constraint about the same end of occurrences eliminates 4 additional models:

\begin{center}
\begin{tabular}{cccc}
$\cancel{(1,2,1,2)}$ & $\cancel{(1,4,1,2)}$ & $\cancel{(1,2,1,4)}$ & $\cancel{(1,4,1,4)}$ \\
$\cancel{(3,2,1,2)}$ & $(3,4,1,2)$ & $(3,2,1,4)$ & $\cancel{(3,4,1,4)}$ \\
$\cancel{(1,2,3,2)}$ & $(1,4,3,2)$ & $(1,2,3,4)$ & $\cancel{(1,4,3,4)}$ \\
$\cancel{(3,2,3,2)}$ & $\cancel{(3,4,3,2)}$ & $\cancel{(3,2,3,4)}$ & $\cancel{(3,4,3,4)}$ \\
\end{tabular}
\end{center}

The constraint about the sequentiality of occurrences eliminates 2 additional models:

\begin{center}
\begin{tabular}{cccc}
$\cancel{(1,2,1,2)}$ & $\cancel{(1,4,1,2)}$ & $\cancel{(1,2,1,4)}$ & $\cancel{(1,4,1,4)}$ \\
$\cancel{(3,2,1,2)}$ & $(3,4,1,2)$ & $\cancel{(3,2,1,4)}$ & $\cancel{(3,4,1,4)}$ \\
$\cancel{(1,2,3,2)}$ & $\cancel{(1,4,3,2)}$ & $(1,2,3,4)$ & $\cancel{(1,4,3,4)}$ \\
$\cancel{(3,2,3,2)}$ & $\cancel{(3,4,3,2)}$ & $\cancel{(3,2,3,4)}$ & $\cancel{(3,4,3,4)}$ \\
\end{tabular}
\end{center}

The final constraint (line 19) eliminates the model $(3,4,1,2)$ which is a symmetric solution of the remaining model $(1,2,3,4)$.

In this case, the resolution constructs a single answer set: \{\lstinline!pattern(1,1). pattern(2,2). occ(1,1,1). occ(1,2,2). occ(2,1,3). occ(2,2,4).!\}.

\espace

This example gives an illustration of successive constraint propagations. This illustration does not correspond to the actual resolution method that uses constraint satisfaction on the grounded program (an intermediary boolean representation of logical rules and facts in the program).

\subsection{Extracting all frequent patterns}
So far, we have seen how to generate patterns of a given length. To extract all the patterns, the program will proceed level-wise:  
generate patterns of length 1, then patterns of length 2, \dots, then patterns of length \lstinline!n!, then patterns of length \lstinline!n+1!, etc. To this end, we use the control facilities introduced in recent releases of \texttt{clingo}. The control part of an ASP program is specified by a python program 
whose main parts are given below. The program begins with an initial grounding step which is followed by successive resolution steps. At each resolution step, the argument value of the \lstinline!patlen(k)! atom is incremented to solve the problem for patterns of size \lstinline!k!.

\begin{lstlisting}[language=Python, numbers=none]
prg.ground("base",[])
for k in range(1,maxsize):
	prg.assignExternal(Fun("patlen", [k]), True)
	prg.solve()
	prg.releaseExternal(Fun("patlen", [k]))
\end{lstlisting}


\subsection{Discussion}

One of the interest of this program is that it generates only a part of all the occurrences of a pattern in a single AS. The program can be seen as an efficient approach for extracting very frequent patterns, \ie $supp(P)\gg \sigma$ (small patterns for instance). We break down (a part of) the algorithmic complexity by extracting only the number of pattern occurrences required by the support threshold:  browsing the whole dataset is not required.
Nonetheless, practically, all the occurrences are actually generated but in different answer sets. If a pattern is very frequent, it would be frequent several times with different subsets of its occurrences. The efficiency to compute an AS is lost by the large number of AS generated within this solution.

In addition, the solution does not enforce the anti-monotonicity property since patterns of size $n$ are computed independently of patterns of size $m<n$.

\espace

To improve this first solution, we explore the incremental resolution facilities of  \texttt{clingo}.

\section{An incremental ASP program for extracting frequent patterns} \label{sec:miningasp2}

\subsection{General principle of incremental extraction of frequent patterns}

We use the incremental control facilities of ASP to implement a level-wise extraction of the frequent patterns: patterns of size $n$ are computed from the patterns of size $n-1$.

\subsubsection{Frequent pattern incremental discovery}

Let $P$ be a frequent pattern of length $n$ ($n\geq 2$) and $\mathcal{I}(P)$ the set of its occurrences. The ``incremental'' discovery of frequent patterns is based on browsing the pattern space level-wise: for each item $s$ in the vocabulary, we would like to know if $Q=P\oplus r$ is a frequent pattern ($Q$ is a right extension of $P$).
The most difficult part of this problem is to build the occurrences of $Q$ incrementally from the occurrences of $P$. 
The following property will be useful to solve this problem: any valid (minimal) occurrence of the extension $Q$ of a pattern $P$ can be obtained by extending some occurrence of pattern $P$ and only one such occurrence can be extended (other extensions will not be minimal).  



\begin{property}\label{prop:augmentation}
Let $n>1$ and $Q=P\oplus r=\langle p_1, \ldots p_{n}, r \rangle $ be a pattern of size $n+1$ extending $P$ and let $S=(s_k)$ be a long sequence., we have:

$\forall I=(i_1, \ldots, i_{n}) \in \mathcal{I}_S(P),\; ( \exists i', s_{i'}=r \wedge \nexists J, J=(j_1, \ldots, j_{n+1}) \in \mathcal{I}_S(Q)$, $[j_1, j_{n+1}]\subset[i_1, i'] \Rightarrow I'=(i_1, \ldots, i_{n}, i') \in \mathcal{I}_S(Q) )$.
\end{property}

%
%

\subsubsection{Incremental ASP program}

The program will be organized in three main parts:
\begin{enumerate}
\item the \lstinline!base! part of the program generates all frequent singleton-patterns (\ie patterns of length 1),
\item the \lstinline!incr(n)! part of the program generates all frequent patterns of length $n$ from patterns of length $n-1$ (extension),
\item the control part defines the global strategy.
\end{enumerate}

To benefit from the backward anti-monotonicity of frequent pattern mining, items are added at the end of a pattern.
The set of occurrences of some pattern can be computed incrementally thanks to property \ref{prop:augmentation}. Contrasting with the previous approach, the complete list of occurrences of the pattern is computed because any occurrence of length $n$ could be useful to compute an occurrence of length $n+1$.


\begin{lstlisting}[language=Python, numbers=none]
prg.ground("base",[])
prg.solve()

for n in range(2,maxsize):
	prg.ground("incr",[n])
	prg.solve()
\end{lstlisting}

This program instantiates the constraints incrementally and solves a partial problem at each step. Moreover, the resolution of \lstinline!incr(n+1)! depends on the AS solved from \lstinline!incr(n)!.

\subsection{Generating patterns of length 1}
The \lstinline!base! part of the incremental program is given below.

\begin{lstlisting}
%generate all frequent symbols between 1 and nbs
symb(S) :- th { seq(P,S) }, S=1..nbs.

% generate patterns of length 1
1{ pattern(1, P) : symb(P) } 1.

% generate all first elements of occurrences 
occ(P, 1, P) :- pattern(1,S), seq(P, S).
\end{lstlisting}

Line 2 generates the list of the frequent symbols. The symbols identifiers are supposed to be between 1 and \lstinline!nbs!. \lstinline!nbs! is a program constant.
Line 5 generates the singleton-patterns, consisting of a single symbol, and line 8 generates the occurrences. This rule enforces to have all the occurrences in the pattern.

 Note that the identifier of an occurrence is the timestamp of the first item (see line 8).



There is no frequency constraints on \lstinline!occ! atoms because this kind of constraint is satisfied by rule in line 2 which selects frequent  symbols only.

\subsection{Generating patterns of length $n$ from patterns of length $n-1$}

Extending patterns means to add an item at the end of a (frequent) pattern of length $n-1$ previously extracted.
The generation of occurrences is based on Property \ref{prop:augmentation}. To obtain the occurrences of an extended pattern, the rule in line 15 attempts to complete each occurrence of the sub-pattern by an \lstinline!occ(I,n,p)! atom. All combinations of \lstinline!occ! atoms that associate the last item of the pattern with a timestamp of the sequence (containing the item) are generated.

We assess, line 18 to 20, that generated occurrences are minimals, and thus we eliminate a large number of AS. 

\begin{lstlisting}[label=asp:IncrFirst, firstnumber=9, caption=Incremental part of the ASP pattern discovery program]
#program incr(n).

% pattern extension
1{ pattern(n, P) : symb(P) } 1.

% occurrences extension
0{ occ(I,n,Q) : seq(Q,S), pattern(n,S), Q>P } 1 :- occ(I, n-1, P).

% minimal occurrences constraints
:- occ(I, n-1, P), occ(I, n, Q), seq(PP,S), pattern(n,S), P<PP, PP<Q.
:- occ(I, n, P), occ(J, n, P), I<J.
:- occ(I, 1, P), occ(J, 1, PP), occ(I, n, Q), occ(J, n, QQ), P<PP, QQ<Q, I!=J.

% frequency constraint
:- { occ(I,n,_) } th-1.
\end{lstlisting}

%
%
%
%
%
%

\subsection{Discussion}

The main interest of this solution is the ability to use the anti-monotonicity property to prune the search space and to a priori avoid the evaluation of many models. The incremental program is correct and complete (due to space limitation, the proof is omitted).

This second program shows that, in ASP, it is simple to generate incrementally the candidate patterns of size $n$ from frequent patterns of size $n-1$. The main difficulty was to rely on pattern occurrences of size $n-1$ to efficiently compute the pattern occurrences of size $n$.  This is achieved thanks to Property \ref{prop:augmentation}.
This ASP program remains simple (only 9 useful lines) and intuitive: the incremental part contains 2 rules, one for extending patterns, one for extending occurrences and 4 constraints for verifying the minimality of occurrences and minimum frequency. 

Contrarily to the first program, all the occurrences of a pattern are collected in a single model. The evaluation of a model is a bit more complex due to more occurrences, but there is much less models to evaluate.

All possible extensions of an occurrence are generated but only the lowest position will generate a minimal occurrence. 
A more efficient way would be to use directive \lstinline!#min! to select the correct occurrence extension or to order possible extensions and select the minimal ones. 



\section{Adding sequential patterns constraints}\label{sec:aspconstraints}

In this section, we show how to include some popular constraints on sequential patterns, \eg \textit{max occurrence duration}, \textit{mingap}, \textit{maxgap}, referring to global constant values.
Such constant values may be defined by \lstinline!#const! directives or be specified in the control part of the program (python part). 

\begin{lstlisting}[numbers=none]
#const maxlength = 20. % maximum duration of an occurrence 
#const maxgap = 7. % maximum delay between two items in an occurrence
#const mingap = 0. % minimum delay between two items in an occurrence
\end{lstlisting}

The generation rules of program \ref{asp:IncrFirst} are modified accordingly. It is better to modify the generation rules than to modify the constraint rules. In the first case, unsatisfiable models are not generated a priori. In the second case, the search space related to the AS is pruned according to  the result of evaluating constraint rules.

In the incremental part of the program, the generation of \lstinline!occ(I, n, Q)! atoms is modified as follows:
\begin{lstlisting}[numbers=none]
0{ occ(I,n,Q) : seq(Q,S), pattern(n,S), Q<=maxgap+P, Q>mingap+P, Q<=maxlength+P } 1 :- occ(I, n-1, P).
\end{lstlisting}

\lstinline!Q! is a valid timestamp for a pattern extension occurrence if it stands within the limits defined by the constraints:
\begin{itemize}
\item gap constraints: $Q \in [mingap+P,maxgap+P]$, where $P$ is the timestamp of the first item
\item duration constraint: $Q \leq maxlength+P$, where $P$ is the timestamp of the first item
\end{itemize}

%
%

\section{Evaluation}

\subsection{Experimentations details}

The ASP programs presented above have been evaluated on computation time. All ASP programs were run using the \texttt{clingo} solver (version 4.3). The \texttt{clingo} processes were limited to a memory space of up to 6 GB. The ASP programs were also compared to the dedicated sequential pattern mining algorithm MinEpi \cite{Mannila1997} through the implementation provided by the tool DMT4SP \cite{Rigotti}. As mentioned in the state of the art, we could not find another generic sequential pattern approach to which we could compare the proposed ASP programs.

The dataset were generated by a random itemset sequence generator. The size of itemsets in the sequence was reduced to 1 and the items were equiprobable. $ql$ denotes the number of different items in the generated sequence.

\subsection{Incremental \textit{vs} non-incremental version}

We briefly compare the non-incremental version with the incremental version on short sequence (20 items). In fact, the non-incremental version quickly overflows the memory.

\begin{table}[ht]
\centering
\begin{tabular}{p{1.9cm}p{2.4cm}p{2.4cm}}
\hline
\textbf{Method} & \textbf{Computation time (sec.)} & \textbf{Max Memory (bytes)}\\ \hline
Non-Incremental &27.05 $\pm$ 5.45& 292996 $\pm$ 916 \\ \hline
Incremental & 0.84 $\pm$ 0.45 & 81336 $\pm$ 10221 \\ \hline
\end{tabular}
\end{table}

\subsection{ASP resolution \textit{vs} dedicated algorithm}

In this section, we compare the incremental ASP program and the DMT4SP implementation of MinEpi. For both programs, the frequency threshold is set to 10\%, the number of different items is set to $ql=10$ and the maximal pattern length is set to 10. For each configuration, the resolution was repeated several times with different sequences. If an execution requires more than the maximal amount of memory allowed then it is not taken into account in the final evaluation.

\begin{figure}[htp]
\centering
\includegraphics[width=.49\textwidth]{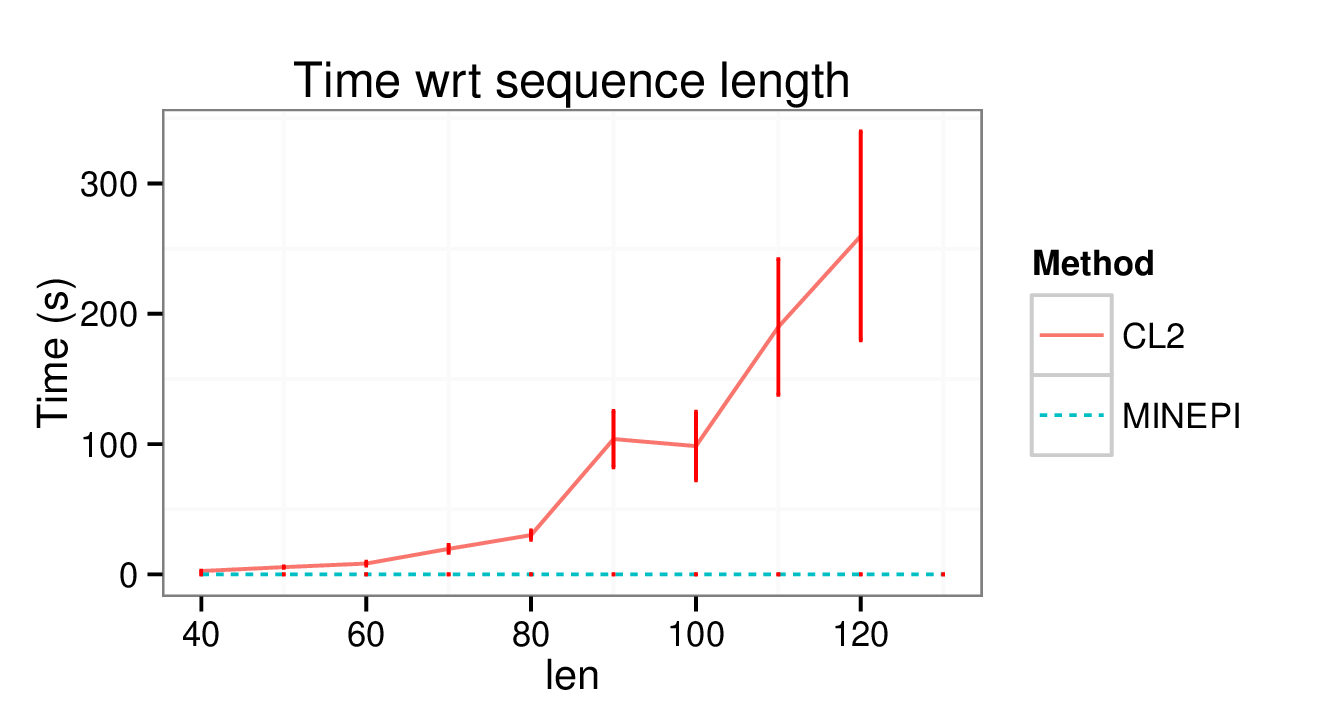} \\
\includegraphics[width=.49\textwidth]{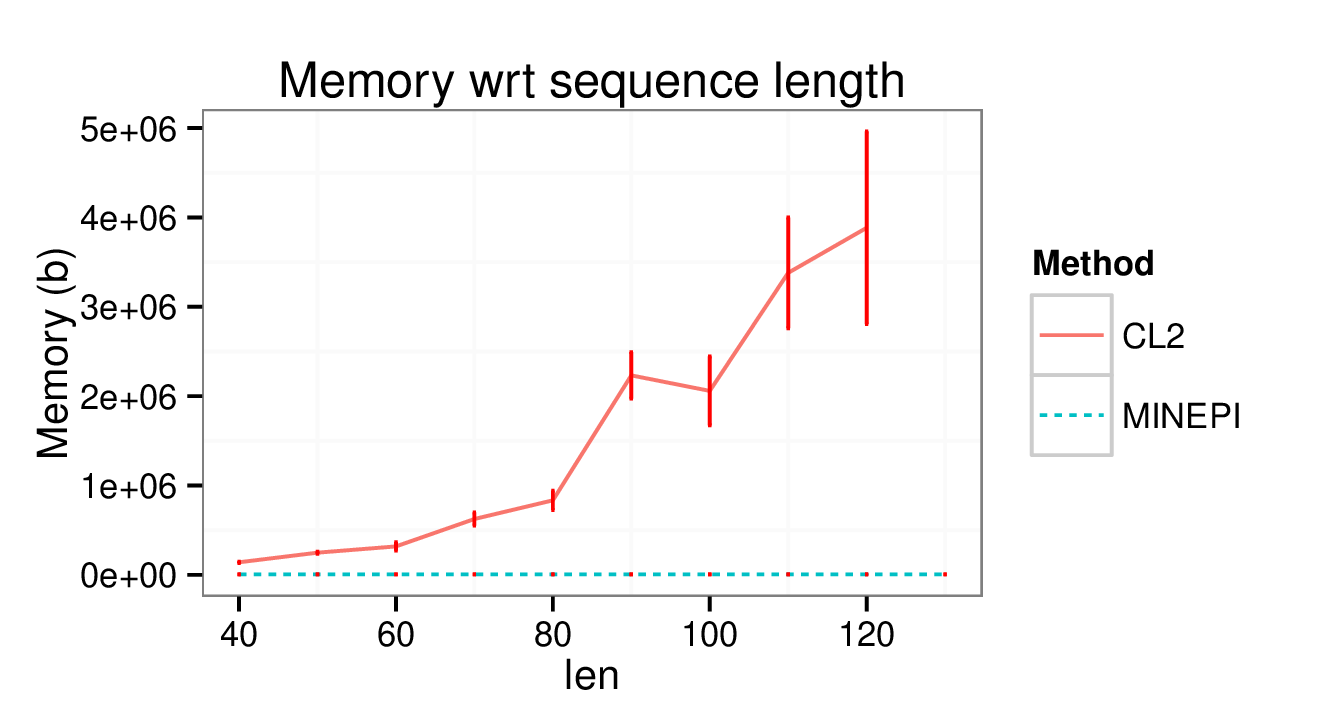}
\caption{Computation time (in seconds) and memory usage (in bytes) of the incremental ASP program wrt to the length of the sequence.}
\label{fig:incrementals}
\end{figure}

Figure \ref{fig:incrementals} illustrates the computation time and memory usage for the ASP program. The computation time increases exponentially with length.  We can note that the memory allocation is strongly related to the computation time. Both time and memory of ASP are several orders of magnitude higher than MinEpi. The computation time by MinEpi is constantly less than 0.01 second for all the sequences of length fewer than 130 items.


\subsection{Computing efficiency improvement ``with constraints''}
In this experiment, we first compared the two incremental ASP programs, without constraints (see section \ref{sec:miningasp2}) and  with sequential pattern constraints (see section \ref{sec:aspconstraints}).
The frequency threshold was still set to 10\% but the size of the vocabulary was set to $ql=7$ (to augment the number of frequent patterns).
The constraints were those defined in the program of section \ref{sec:aspconstraints}.

Adding constraints reduces the combinatorics of the search space and, especially, the combinatorics of occurrences extensions. As a consequence, the number of generated models is considerably reduced as well as the computation time.
For instance, the mean computation time for a sequence of 70 items is 166\textit{s} $\pm$ 34\textit{s} without constraints while with constraints it is 37\textit{s} $\pm$ 27\textit{s}. Using constraints, the ASP program becomes computationally more competitive with the dedicated algorithm.

\begin{figure}[htp]
\centering
\includegraphics[width=.45\textwidth]{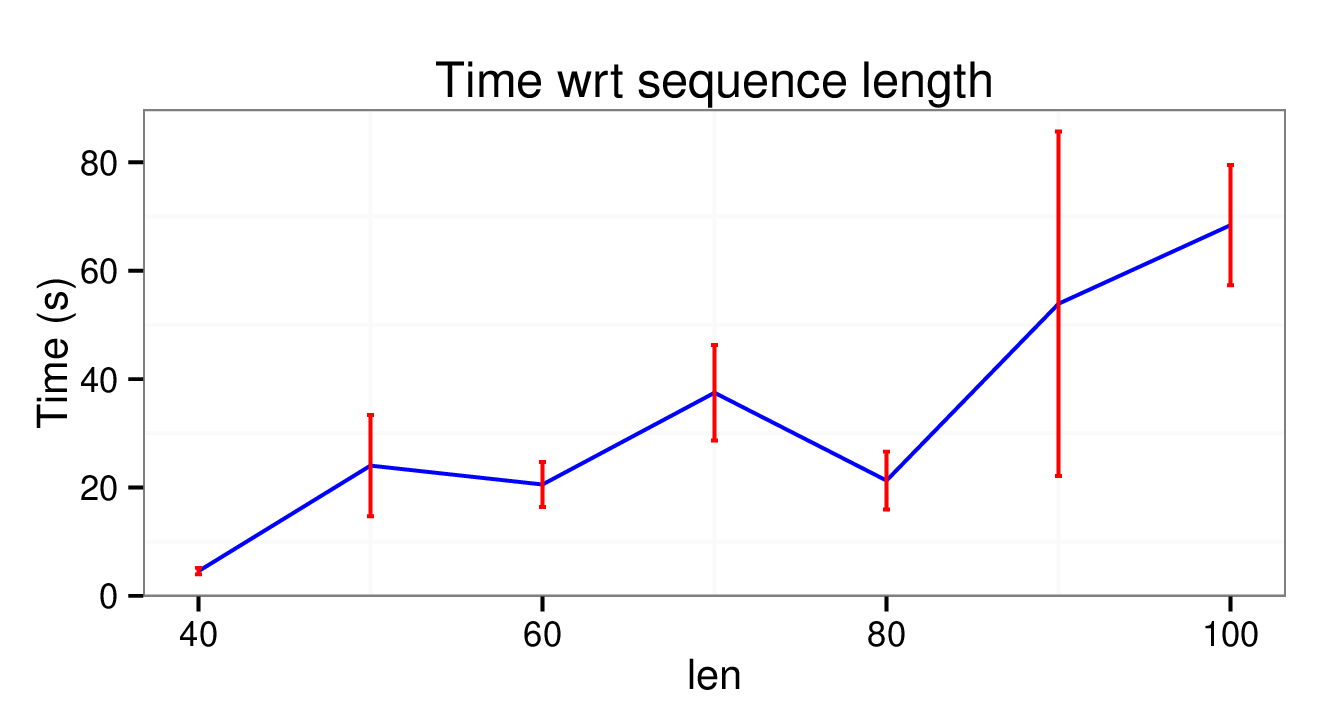}
\caption{Computation time (in seconds) of the incremental ASP program ``with constraints'' wrt to the length of the sequence. }
\label{fig:constraints1}
\end{figure}

Figure \ref{fig:constraints1} illustrates the computation time for the ASP program. The computation times are lower than those of the figure \ref{fig:incrementals}. Nonetheless, computation time still increases exponentially with the length of the sequence.

\section{Conclusion}
To the best of our knowledge, this is the first proposed ASP program for frequent pattern mining that implements the anti-monotonicity constraint to prune the search space.
It takes advantage of the control facilities of a recent release of \texttt{clingo} \cite{gekakaosscsc11a} to design an incremental ASP program.
The results show that our proposal is very slow compared to dedicated algorithms such as MinEpi, but our main objective was to demonstrate the feasability of a declarative and generic approach. ASP solutions are known to be competitive on hard problems. 
Maybe, enumerating occurrences of a string pattern is not sufficiently hard to be competitive with dedicated algorithms. To assess this assumption, we are working on mining frequent episodes (sequential patterns as sequences of itemsets).

There is room for improving our solution. An immediate improvement would be to use an ordered list of possible extensions.
Another classical challenge in pattern mining would be for ASP to tackle the extraction of closed patterns which can considerably improve the performance of data mining algorithms. The closure definition involves evaluating several patterns together, so the current approach (one model = one pattern) does not seem to be relevant to cope with such constraints. The use of the \lstinline!#external! directive could be explored to achieve this goal.

Another perspective is to investigate how this framework can be extended to deal with other classical pattern mining problems such as tree mining, subgraph mining, etc.
Finally, we would like to explore how to let the user express his own constraints on patterns. A meta-language that could automatically generate constraints in ASP is a midterm objective.

\bibliographystyle{plain}
\bibliography{biblio}

\end{document}